\pdfoutput=1
\documentclass[pmlr]{jmlr}

\usepackage{longtable}% for long tables

\usepackage{booktabs}
\usepackage[load-configurations=version-1]{siunitx} % newer version
 \usepackage{enumitem}

 \newcommand{\fix}[1]{#1}

\jmlrvolume{1}
\jmlryear{2018}
\jmlrworkshop{Learning with Imbalanced Domains: Theory and Applications}

\title[Learning from Positive and Unlabeled Data under the SAR Assumption]{Learning from Positive and Unlabeled Data \\ under the Selected At Random Assumption}

  \author{\Name{Jessa Bekker} \Email{jessa.bekker@cs.kuleuven.be}\\ 
   \Name{Jesse Davis} \Email{jesse.davis@cs.kuleuven.be}\\
   \addr KU Leuven, Belgium}

\editor{}
\begin{document}

\maketitle

\begin{abstract}
For many interesting tasks, such as medical diagnosis and web page classification, a learner only has access to some positively labeled examples and many unlabeled examples. Learning from this type of data requires making assumptions about the true distribution of the classes and/or the mechanism that was used to select the positive examples to be labeled. The commonly made assumptions, separability of the classes and positive examples being selected completely at random, are very strong. This paper proposes a weaker assumption that assumes the positive examples to be selected at random, conditioned on some of the attributes. To learn under this assumption, an EM method is proposed. Experiments show that our method is not only very capable of learning under this assumption, but it also outperforms the state of the art for learning under the selected completely at random assumption.
\end{abstract}

\section{Introduction}
%binary classifier -> pu setting
When learning binary classifiers from fully labeled data, algorithms have access to the class labels for all examples. However, in practice, many data sets only provide positive labels for some examples, with the remaining data being unlabeled and containing both positive and negative examples. Learning from positive and unlabeled data (PU learning) attempts to learn a classifier from this data. \fix{PU learning is closely related to semi-supervised learning and one-class classification \citep{khan_madden_2014}.}

%example domains
The problem of positive and unlabeled data arises often in practice. Medical records, for example, list the diseases that patients have been diagnosed with. However, some diseases, like diabetes, often go undiagnosed.
 In this case, it is wrong to assume that an undiagnosed patient does not have the disease. Another examples are bookmarked pages, as these are but a subset of the pages of interest \citep{lee2003learning,liu2003building}, and knowledge bases, which only contain a subset of facts \citep{zupanc2018estimating}.

%approaches: separability or scar
 \fix{There are roughly three established assumptions that enable  PU learning. 1) Assuming the unlabeled data to be negative \citep{neelakantan2015compositional}. 2) Assuming separability, i.e., that the negative examples are very different from the positive  ones. Learning then consists of two steps: finding reliable negative examples and then applying standard machine learning \citep{li2003learning,yu2005single,nguyen2011positive}. 3) Assuming that the labeled examples are selected completely at random from the set of positive examples, a classifier can be learned by incorporating the probability of labeling an positive example \citep{denis1998pac,lee2003learning,liu2003building,liu2005partially,denis2005learning,zhang2005simple,elkan2008learning,mordelet2014bagging,claesen2015robust}.}

%this paper: SAR
The aforementioned assumptions are very strong. Our goal in this paper is to enable learning under weaker assumptions by introducing a new assumption to enable learning from positive and unlabeled data: the {\it Selected At Random (SAR)} assumption. It is related to the third category of approaches, but instead of assuming a constant probability for all positive examples to be labeled, it assumes that the probability is a function of the attributes, called the propensity score. The propensity score originates from causal inference, but has also been applied in semi-supervised learning \citep{causalInference,Schnabel2016RecommendationsAT}. In order for it to be possible to learn in this setting, our proposed method SAR-EM assumes that the propensity score only depends on a known subset of the attributes.

%contributions
Our contributions are the following: 1) formulate the SAR assumption for learning from positive and unlabeled data, 2) propose a method SAR-EM that works under this assumption, 3) show that a special case of SAR-EM  which assumes the SCAR assumption outperforms the state-of-the-art methods for estimating the class prior in PU data, 4) show that incorrectly making the SCAR assumption hurts the classifier performance, and 5) show that SAR-EM can reconstruct propensity score functions and learn good classifiers.

\section{Background}

% What is pu learning + notation
The goal of learning from positive and unlabeled (PU) data is to train a binary classifier while only having access to positively labeled and unlabeled examples.  An example is represented by $\{x,y,s\}$, where $x$ are its attributes, $y$ the true class and $s$ the label. Only positive examples are labeled: $s=1 \Rightarrow y=1$, and unlabeled examples $s=0$ can be of any class $y$. Bold letters depict sets of variables, i.e., the dataset is represented by $\{\mathbf{x},\mathbf{y},\mathbf{s}\}$.

% Assumptions

To enable learning with positive and unlabeled data, assumptions about the population of positives and negatives in the instance space are necessary. The two most popular assumptions are 1) separability and 2) selected completely at random. The separability assumption assumes that within the considered class of models a model exists that can perfectly separate the positive from the negative examples. This assumption is violated if the attributes do not contain enough information to deterministically determine the class. When separability holds, two-step approaches can be used to solve the problem \citep{li2003learning,yu2005single,li2009positive,nguyen2011positive}.

Learning without the separability assumption is challenging because examples with a low probability of being positive could still have a positive label. In this case, the sampling mechanism to select positive examples to be labeled needs to be considered. The most common assumption is that the labeled positives are a random subsample of the true positives. This is the selected completely at random assumption.

\subsection{Selected Completely at Random Assumption}
Under the {\it Selected Completely At Random (SCAR)} assumption, every positive example has exactly the same probability to be selected to be labeled \citep{elkan2008learning}. Given the class, this probability (the {\it label frequency} $c$), is conditionally independent of the attributes:

\begin{equation}
\label{eq:iid}
c = \Pr(s=1|\mathbf{x},y=1) = \Pr(s=1|y=1).
\end{equation}

The SCAR assumption was introduced in analogy with the {\it Missing Completely A Random assumption (MCAR)} that is common when working with missing data \citep{rubin1976inference,little2014statistical}. However, there is a notable difference between the two assumptions. In MCAR data, the missingness of the variable cannot depend on the value of the variable, where in PU learning this is necessarily the case because all negative labels are missing. The class values are missing completely at random only if just the population of positive examples is considered.

A very useful property follows from the SCAR assumption: An example's probability of belonging to the positive class is directly proportional to the probability of that example being labeled \citep{elkan2008learning}:
 \begin{equation}
 \label{eq:ycs}
 \Pr(y=1|\mathbf{x}) = \frac{1}{c}\Pr(s=1|\mathbf{x}).
 \end{equation}

A probabilistic model to predict $\Pr(y=1|\mathbf{x})$ can be obtained by training a model that predicts $\Pr(s=1|\mathbf{x})$ from the data and scaling the output probabilities with $1/c$. This method only works with well-calibrated models and is therefore not always robust. Other methods have been introduced to cope with this, they integrate the label frequency in the training process \citep{denis1998pac,de1999positive,liu2003building,zhang2005simple,elkan2008learning}.

The previously mentioned methods require the label frequency, or equivalently the class prior $\alpha=\Pr(y=1)=\Pr(s=1)/c$, to be known. The class prior can be known from domain knowledge or could be estimated by labeling a small random sample of the dataset.

A substantial effort has been done to estimate the label frequency $c$ directly from the data \citep{elkan2008learning,Plessis2015ClasspriorEF,Jain2016EstimatingTC,Ramaswamy2016MixturePE,Bekker2018AAAI,BekkerILP17}. The common assumption that these methods make is that in some region in the instance space, the positive class probability should be 1. We refer to this assumption as the {\it local certainty assumption}.

\section{Selected at Random Assumption}
\label{sec:sar}
% SAR: What + connections missing data
This paper introduces the notion of positive examples being {\it Selected At Random (SAR)} to be labeled. This assumes that the selection probability is completely determined by its attributes. Just like SCAR has MCAR as a counterpart in the missing data literature, SAR is based on the {\it Missing At Random (MAR)} assumption \citep{rubin1976inference, little2014statistical}.

\subsection{Propensity Score}
The probability that a positive example is selected to be labeled is the {\it propensity score} $e(x)$:
\[
e(x) = \Pr(s=1|y=1,x).
\]
Note that the propensity score only applies to positive examples; negative examples never get selected. This seems to imply that negative examples are never considered to be labeled. However, this is not necessarily the case, the labels could be lost afterwards. For example, doctors test both ill and healthy patients for diseases but may only store positive results in the patients' records. In this case, the propensity score is the unconditional probability of testing a patient and the conditional probability of storing the disease in the record.

\subsection{Learning with a Known Propensity Score}
\label{sec:knowne}
The propensity score is known when the selection mechanism for positive examples is understood. For example, a hospital might have a protocol for testing people.
An unbiased classifier can then be trained from positive and unlabeled data by taking the propensity score into account in an equivalent way as the label frequency under the SCAR assumption: by either scaling the output probabilities with $1/e(x)$ or integration in the training process.

\subsection{Learning with an Unknown Propensity Score}
This paper's goal is to learn from positive and unlabeled data under the SAR assumption with an unknown propensity score. It is an ill-defined problem because any unlabeled example can be explained by both a low positive class probability and a low labeling probability. This section evaluates which reasonable additional assumptions would enable learning.

\paragraph{Propensity Attributes}
The first assumption is that the propensity function only requires a subset of the attributes $x_e \in x$:
\begin{align*}
\Pr(s=1|y=1,x) &= \Pr(s=1|y=1,x_e)\\
e(x)&=e(x_e).
\end{align*}
This is often a reasonable assumption: not all of the attributes might have been available at selection time, some of the attributes might be difficult to interpret for a labeler, or it might be known which variables have the highest correlation with the class and the labeler only used those.

\paragraph{SAR as Multi-SCAR}
Given discrete propensity attributes, the classification problem under the SAR assumption can be reduced to multiple classification problems under the SCAR assumption by partitioning the population into strata based on assignments to values of the propensity attributes. While being suboptimal in practice, this approach is insightful for theoretical analysis. Indeed, the conditions needed for training a correct classifier in each of the strata are sufficient for a correct classifier over the entire population.

\paragraph{Local Certainty Assumption for Any Propensity Configuration}
\label{sec:localcertainty}
To enable learning with an unknown label frequency under the SCAR assumption, {\it local certainty} is commonly assumed.
This means that the probability of examples belonging to the positive class needs to be 1 in some region of the attribute domain. 
A sufficient condition for learning under the SAR assumption to be possible is therefore that {\it local certainty} holds in all the propensity strata. Although seemingly very strong, this assumption is not implausible. It holds, for example, if one of the non-propensity attributes always provides certainty of the positive class. In the task of classifying web pages as commercial, any page with a ``buy'' button on it is commercial, regardless of the tags that the labeler had access to when choosing the pages to label.

We argue that the local certainty assumption for any propensity configuration can be relaxed if the propensity score function and the classification function have a certain smoothness over the propensity attributes. In other words, some similarity between the classification function when conditioned on different propensity configurations is expected. With this insight, we propose to simultaneously train a classifier and a propensity score function while promoting local certainty as much as possible.

\begin{algorithm2e}
	\caption{SAR-EM}
	\label{alg:sarem}
	% older versions of algorithm2e have \dontprintsemicolon instead
	% of the following:
	\DontPrintSemicolon
	%\dontprintsemicolon
	
	% older versions of algorithm2e have \linesnumbered instead of the
	% following:
	\LinesNumbered
	%\linesnumbered
	\KwIn{attributes $\mathbf{x}$, labels $\mathbf{s}$, propensity attributes $\mathbf{x_e}$, local certainty parameter $d$}
	\KwOut{classifier $f$, propensity score model $e$}
	$f,e = \text{initialize\_models}(\mathbf{x}, \mathbf{s}, \mathbf{x_e})$\;
	
	\Repeat{Converged or maximum iterations reached}{
		// Expectation\;
		$\mathbf{\hat y_f} = f(\mathbf{x}[\mathbf{s}==0])$\;
		$\mathbf{\hat s} = d\cdot e(\mathbf{x_e}[\mathbf{s}==0])$ // Decay propensity scores\;
		
		$\mathbf{\hat y} = \frac{\mathbf{\hat y_f}(1-\mathbf{\hat s})}{\mathbf{\hat y_f}(1-\mathbf{\hat s}) + (1-\mathbf{\hat y_f})}$\;
		
		\vspace{5pt}
		// Maximization\;
		$\mathbf{y_w} = \text{ones}(\text{size}(\mathbf{s}))::\text{zeros}(\text{size}(\mathbf{s}[\mathbf{s}==0]))$\;
		$\mathbf{x_w} = \mathbf{x}[\mathbf{s}==1]::\mathbf{x}[\mathbf{s==0}]::\mathbf{x}[\mathbf{s==0}]$\;
		$\mathbf{w} = \mathbf{s}[\mathbf{s}==1]::\mathbf{\hat y}::(1-\mathbf{\hat y})$\;
		$f = \text{fit}(\mathbf{x_w}, \mathbf{y_w}, \mathbf{w})$\;
		\vspace{5pt}
		$\mathbf{s_w} = \mathbf{s}[\mathbf{s}==1] :: \mathbf{s}[\mathbf{s}==0]$\;
		$\mathbf{x_{ew}} = \mathbf{x_e}[\mathbf{s}==1]::\mathbf{x_e}[\mathbf{s==0}]$\;
		$\mathbf{w} = \mathbf{s}[\mathbf{s}==1]::\mathbf{\hat y}$\;
		$e = \text{fit}(\mathbf{x_{ew}}, \mathbf{s_w}, \mathbf{w})$\;
	}
	\Return{$f,e$}\;
	
\end{algorithm2e}

\section{An EM Method for PU Learning under the SAR Assumption}
\label{sec:sarem}
We use an Expectation Maximization (EM) approach to solve the learning problem. The true class values $y$ are hidden variables, while attributes $x$ and labels $s$ are observed and the propensity attributes are known. The data is generated by the following process:
\begin{align*}
(x,y,s)	&\sim \Pr(x,y,s)\\
		&\sim \Pr(x)\Pr(y|x)\Pr(s|x,y)\\
		&\sim \Pr(x)\Pr(y|x)\Pr(s|x_e,y),
\end{align*}
where $x_e$ are the propensity attributes. 
We assume that the process of classifying examples and labeling examples can be modeled using parameters $\theta$ and $\phi$ respectively:
\begin{align*}
(x,y,s)	&\sim \Pr(x)\Pr(y|x,\theta)\Pr(s|x_e,y,\phi).
\end{align*}
The goal is to find the 
parameters $\theta$ and $\phi$ that explain the observed data the best.

\subsection{Expectation Maximization}
Optimizing parameters $\theta$ and $\phi$ means setting them to maximize the log likelihood of observing $(x,s)$. Expectation maximization repeats two steps until the models e. During the {\it expectation} step, it finds the expected values $\hat y$ for $y$ given the current models. During the {\it maximization} step, it retrains the models to optimize the log likelihood of observing $(x,s,\hat y)$. The steps are derived below.

\paragraph{Expectation}  (\algorithmref{alg:sarem}, lines 3-7)
\begin{align*}
&\Pr(y=1|x, s=1, \theta,\phi) =  1 \\\nonumber\\
&\Pr(y=1|x, s=0, \theta,\phi) \\
&\quad= \frac{\Pr(x, s=0|y=1, \theta,\phi)\Pr(y=1|\theta,\phi)}{\Pr(x, s=0|\theta,\phi)}\\
&\quad= \frac{\Pr(x|y=1, \theta,\phi)\Pr(s=0|x, y=1,\theta,\phi)\Pr(y=1|\theta,\phi)}{\Pr(x, s=0|\theta,\phi)}\\
&\quad= \frac{\frac{\Pr(y=1|x,\theta)\Pr(x|\theta,\phi)}{\Pr(y=1|\theta,\phi)}\Pr(s=0|x_e, y=1,\phi)\Pr(y=1|\theta,\phi)}{\Pr(x, s=0|\theta,\phi)}\\
&\quad= \frac{\Pr(y=1|x,\theta)\Pr(x|\theta,\phi)\Pr(s=0|y=1,x_e,\phi)} {\Pr(s=0|x,\theta,\phi)\Pr(x|\theta,\phi)}\\
&\quad= \frac{\Pr(y=1|x,\theta)\Pr(s=0|y=1,x_e,\phi)} {\Pr(s=0|x,\theta,\phi)}\\
&\quad= \frac{\Pr(y=1|x,\theta)\Pr(s=0|y=1,x_e,\phi)} {\Pr(y=1|x,\theta)\Pr(s=0|y=1,x,\phi)+\Pr(y=0|x,\theta)\Pr(s=0|y=0,x,\phi)}\\
&\quad= \frac{\Pr(y=1|x,\theta)\Pr(s=0|y=1,x_e,\phi)} {\Pr(y=1|x,\theta)\Pr(s=0|y=1,x_e,\phi)+\Pr(y=0|x,\theta)}
\end{align*}

\paragraph{Maximization}(\algorithmref{alg:sarem}, lines 8-16)
\begin{align*}
\max_{\theta,\phi}\mathbb{E}_{y|x,s,\theta,\phi}\log\Pr(x, s, y|\theta,\phi)\nonumber & =\max_{\theta,\phi}\mathbb{E}_{y|x,s,\theta,\phi}\log\left[\Pr(x)\Pr(y|x,\theta)\Pr(s|y,x_e,\phi)\right]\nonumber\\
&=\max_{\theta,\phi}\mathbb{E}_{y|x,s,\theta,\phi}\log\left[\Pr(y|x,\theta)\Pr(s|y,x_e,\phi)\right]\nonumber\\
&=\max_{\theta,\phi}\mathbb{E}_{y|x,s,\theta,\phi}\left[\log\Pr(y|x,\theta)+\log \Pr(s|y,x_e,\phi)\right]\nonumber\\
%&=\max_{\theta,\phi}\left[\mathbb{E}_{y|x,s,\theta,\phi} \log\Pr(y|x,\theta)+\mathbb{E}_{y|x,s,\theta,\phi}\log \Pr(s|y,x_e,\phi)\right]\nonumber\\
&=\max_{\theta}\mathbb{E}_{y|x,s,\theta,\phi} \log\Pr(y|x,\theta)+\max_{\phi}\mathbb{E}_{y|x,s,\theta,\phi}\log \Pr(s|y,x_e,\phi)\nonumber\\
%&=\max_{\theta}\mathbb{E}_{y|x,s,\theta,\phi} \log\Pr(y|x,\theta)+\max_{\phi}\Pr(y=1|x,s,\theta,\phi)\log \Pr(s|y=1,x_e,\phi)\nonumber
\end{align*}

\begin{algorithm2e}
	\caption{initialize\_models}
	\label{alg:init}
	%\linesnumbered
	%\dontprintsemicolon
	\LinesNumbered
	\DontPrintSemicolon
	
	\KwIn{attributes $\mathbf{x}$, labels $\mathbf{s}$, propensity attributes $\mathbf{x_e}$}
	\KwOut{classifier $f$, propensity score model $e$}
	// Fit $f$ to predict $s$ from $x$\;
	
	$f = \text{fit}(\mathbf{x},\mathbf{s})$\;
	\vspace{5pt}
	// SCAR assumption with minimum $c$ \;
	$\mathbf{\hat s} = f(\mathbf{x}[\mathbf{s}==0])$\;
	$c = 1/\max(\mathbf{\hat s})$ \;
	
	\vspace{5pt}
	// Fit $f$ to predict $s$ from $x$ with $c$ \;
	
	$\mathbf{\hat y} = \frac{1-c}{c}\frac{\hat s}{1-\hat s}$\;

	$\mathbf{y_w} = \text{ones}(\text{size}(\mathbf{s}))::\text{zeros}(\text{size}(\mathbf{s}[\mathbf{s}==0]))$\;
	$\mathbf{x_w} = \mathbf{x}[\mathbf{s}==1]::\mathbf{x}[\mathbf{s==0}]::\mathbf{x}[\mathbf{s==0}]$\;
	$\mathbf{w} = \mathbf{s}[\mathbf{s}==1]::\mathbf{\hat y}::(1-\mathbf{\hat y})$\;
	$f = \text{fit}(\mathbf{x_w}, \mathbf{y_w}, \mathbf{w})$\;
	
	\vspace{5pt}
	
	// Train $e$ to return $c$ for any input	
	
	$\mathbf{s_w} = \text{ones}(\text{size}(\mathbf{s}))::\text{zeros}(\text{size}(\mathbf{s}))$\;
	$\mathbf{x_{ew}} = \mathbf{x_{e}}::\mathbf{x_{e}}$\;
	$\mathbf{w} = c\cdot\text{ones}(\text{size}(\mathbf{s}))::(1-c)\cdot\text{ones}(\text{size}(\mathbf{s})$\;
	$e = \text{fit}(\mathbf{x_{ew}}, \mathbf{s_w}, \mathbf{w})$ \;
	\vspace{5pt}
	
	\Return{f,e}\;
	
\end{algorithm2e}

\subsection{{Local Certainty}}
Applying pure EM ensures converging to a combination of classifier and propensity score model that explains the observed data well. However, care must be taken because a propensity score model that always returns 1 and a classifier that predicts the probability of observing a label explains the observed data perfectly but is not the desired solution. 
As stated in~\sectionref{sec:localcertainty}, classifiers with local certainty are preferred, so this needs to be taken into account during the EM optimization process.

To guide the learning towards local certainty, lower propensity scores need to be encouraged. 
To this end, the predicted propensity scores can simply be decayed at every iteration during the expectation step (\algorithmref{alg:sarem}, line 6). This enhances the expected class probabilities and makes sure that a more positively inclined classifier is trained.

\subsection{Initialization}
To start EM with reasonable models, a classifier and propensity model are trained under the SCAR assumption. The label frequency is estimated by training a model to predict $\Pr(s=1|x)$ and setting the label frequency so that the maximum predicted class probability becomes 1 (\algorithmref{alg:init}, lines 1-5). This is estimator 3 of \citet{elkan2008learning}, which gives a fairly unstable estimate, but it is fine for initialization.

Subsequently, the classifier is trained to predict $y$ by weighting the examples using the label frequency 
\citep{elkan2008learning}
(\algorithmref{alg:init}, lines 6-11). The propensity score model is trained to always return $c$, by providing all examples as both positive and negative, but giving weight $c$ when positive and $1-c$ otherwise (\algorithmref{alg:init}, lines 12-16).

\subsection{Convergence}
Convergence is assumed when the propensity score predictions do not change much over several iterations. Change is quantified as the average absolute slope of the minimum mean square error line through the predictions:

\begin{equation*}
\text{slope}(s,t,n) = \frac{n \sum_{i=0}^{n-1} (i s_{t-n+1+i}) - \sum_{i=0}^{n-1}i \sum_{i=0}^{n-1}s_{t-i} }{n \sum_{i=0}^{n-1}i^2 - (\sum_{i=0}^{n-1}i)^2}
\end{equation*}
\begin{equation*}
\frac{1}{|\mathbf{\hat s}|}\sum_{\hat s \in \mathbf{\hat s}}|\text{slope}(s,t,n)| <\epsilon,
\end{equation*}
where $t$ is the current iteration, $n$ the number of iterations over which the slope is taken, $s_i$ the propensity score prediction during iteration $i$, and $\epsilon$ the minimum average absolute slope for non-convergence. Additionally, a maximum number of iterations can be set.

\subsection{SCAR with Label Frequency Estimation as a Special Case of SAR-EM}
\label{sec:sarscar}
SAR-EM can be used to train a classifier and estimate the label frequency under the SCAR assumption, by assigning no propensity attributes: $x_e=\emptyset$. Training the propensity score model then reduces to estimating the label frequency $c$ given the expected class values.

\subsection{Learning with a Known Propensity Score as a Special Case of SAR-EM}
\label{sec:sar-e}
If the propensity score (or label frequency) is known, SAR-EM can still optimize the classifier. The algorithm remains the same, except for not training the propensity score model, i.e., removing lines 13-16 in \algorithmref{alg:sarem}  and 12-16 in \algorithmref{alg:init}.
	
Training the classifier using SAR-EM is more stable than predicting the labels and adjusting the output probabilities (\sectionref{sec:knowne}). By optimizing the expected likelihood of the observed data, the influence of unexpectedly labeled low probability positives is reduced.

\begin{table}[tbp]
	\floatconts
	{tab:data}%
	{\caption{Datasets}}%
	{\small\begin{tabular}{l|rrrr}
			Dataset	& \# Examples	&\# Vars & $\Pr(y=1)$\\\hline
			Breast Cancer	& 683		& 9 	& 0.350\\
			Mushroom	& 8,124		& 21 	& 0.482 \\
			Adult		& 48,842	& 14  	& 0.761\\
			IJCNN		& 141,691	& 22  	& 0.096\\
			Cover Type	& 536,301	& 54  	& 0.495\\
			20ng Comp - Rec	&	5,287	& 200	& 0.450\\
			20ng Comp - Sci	&	5,279	& 200	& 0.450\\
			20ng comp - Talk	&	4,856	& 200	& 0.401\\
			20ng rec - Sci	&	4,752	& 200	& 0.499\\
			20ng Rec - Talk	&	4,329	& 200	& 0.450\\
			20ng Sci - Talk	&	4,321	& 200	& 0.451
		\end{tabular}}
	\end{table}
	
	\begin{table}[tbp]
		\floatconts
		{tab:c}%
		{\caption{Class prior estimation: SAR-EM has the best rank and lowest absolute error.}}%
		{\small
			\begin{tabular}{l|cc}
				Method	&	Average $|\hat \alpha - \alpha|$ rank +/- SD	&	Average $|\hat \alpha - \alpha|$ +/- SD	\\\hline
				SAR-EM	&2.87 +/- 1.83	&0.08 +/- 0.07\\
				KM2	&3.38 +/- 2.38	&0.10 +/- 0.13\\
				TI$c$E	&3.86 +/- 2.13	&0.09 +/- 0.06\\
				AlphaMax\_N	&4.04 +/- 2.15	&0.13 +/- 0.10\\
				AlphaMax	&4.16 +/- 1.72	&0.12 +/- 0.09\\
				KM1	&4.92 +/- 2.01	&0.11 +/- 0.09\\
				EN	&6.75 +/- 1.78	&0.27 +/- 0.16\\
				PE	&7.02 +/- 1.61	&0.29 +/- 0.14\\
				pen-L1	&7.81 +/- 2.03	&0.37 +/- 0.21\\
			\end{tabular}
		}
	\end{table}

\section{Empirical Evaluation}
\label{sec:exp}
Our experiments aim to evaluate the performance of SAR-EM. Because, to the best of our knowledge, no other methods exist to learn from positive and unlabeled data under the SAR assumption, we will first compare our method under the SCAR assumption to other methods that assume SCAR and estimate the class prior in this data. Next, the SAR assumption is considered. Here, our method is compared to SCAR methods, supervised methods and SAR-EM when the propensity score function is known. \fix{Lastly, we analyze the performance in unbalanced domains.}

\subsection{SAR-EM Settings} Logistic regression is used for both the classifier and the propensity score model because it is a simple classifier that is known to predict well-calibrated probabilities \citep{niculescu2005predicting}. All experiments have local certainty parameter $d$=0.9 and  convergence parameters $n=10$, $\epsilon=0.0001$.

\begin{figure}[tbp]
	\floatconts
	{fig:c}
	{\caption{Class prior estimation comparison}}
	{%
		\hspace{0.05\linewidth}\includegraphics[width=0.14\linewidth]{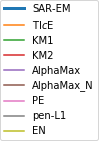}\hspace{0.01\linewidth}\quad
		\includegraphics[width=0.2\linewidth]{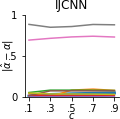}\quad
		\includegraphics[width=0.2\linewidth]{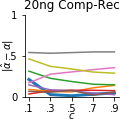}\quad
		\includegraphics[width=0.2\linewidth]{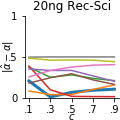}\quad
		\vspace{10pt}
		
		\includegraphics[width=0.2\linewidth]{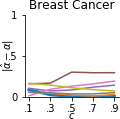}\quad
		\includegraphics[width=0.2\linewidth]{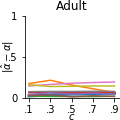}\quad
		\includegraphics[width=0.2\linewidth]{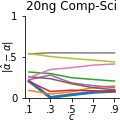}\quad
		\includegraphics[width=0.2\linewidth]{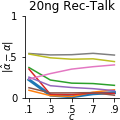}\quad
		\vspace{10pt}
		
		\includegraphics[width=0.2\linewidth]{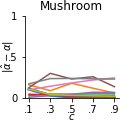}\quad
		\includegraphics[width=0.2\linewidth]{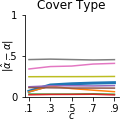}\quad
		\includegraphics[width=0.2\linewidth]{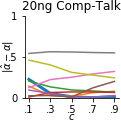}\quad
		\includegraphics[width=0.2\linewidth]{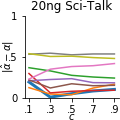}\quad
	}	
\end{figure}

\subsection{Performance under the SCAR Assumption}
To evaluate how well SAR-EM does in the SCAR setting, the same datasets (\tableref{tab:data}) are used as in \citet{Bekker2018AAAI} to benchmark methods for label frequency estimation. \fix{IJCNN originates from the IJCNN 2001 neural network competition.\footnote{Available on: \url{https://www.csie.ntu.edu.tw/~cjlin/libsvmtools/datasets/}} The other dataset are available on the  UCI repository.\footnote{\url{http://archive.ics.uci.edu/ml/}} The preprocessed datasets of \citet{Bekker2018AAAI} are used here, which means that the multivalued features are binarized and numerical features scaled between 0 and 1. To generate binary classification datasets from the twenty newsgroups (20ng), different categories are compared (computer, recreation, science and talk) and the features are generated using bag of words with the 200 most frequent words, disregarding nltk stopwords.}

The following class prior estimation methods are compared: EN \citep{elkan2008learning}, PE \citep{Plessis2014ClassPE}, pen-L1 \citep{Plessis2015ClasspriorEF}, KM1 and KM2 \citep{Ramaswamy2016MixturePE}, AlphaMax \citep{Jain2016NonparametricSL} and AlphaMax\_N \citep{Jain2016EstimatingTC}. The datasets were turned into PU datasets in the same way as done by \citet{Bekker2018AAAI}: the positive examples are selected to be labeled with label frequencies $c\in[0.1,0.3,0.5,0.7,0.9]$.

As usual, the algorithms based on their absolute error in class prior $|\hat\alpha-\alpha|$.

\begin{figure}[tbp]
	\floatconts
	{fig:delta3}
	{\caption{Propensity score with one variable, centered around $\overline{c}=0.3$. Assuming SCAR ($\Delta c=0$) hurts the performance when the reality deviates from it. Knowing the propensity function does not result in a big benefit over estimating it.}}
	{%
		\includegraphics[width=0.18\linewidth]{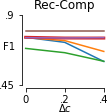}\quad
		\includegraphics[width=0.18\linewidth]{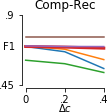}\quad
		\includegraphics[width=0.18\linewidth]{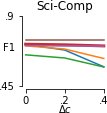}\quad
		\includegraphics[width=0.18\linewidth]{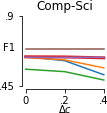}\quad
		\vspace{10pt}
		
		\includegraphics[width=0.18\linewidth]{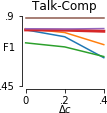}\quad
		\includegraphics[width=0.18\linewidth]{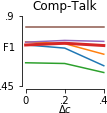}\quad
		\includegraphics[width=0.18\linewidth]{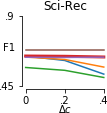}\quad
		\includegraphics[width=0.18\linewidth]{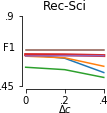}\quad	
		\vspace{10pt}
		
		\includegraphics[width=0.18\linewidth]{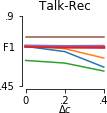}\quad
		\includegraphics[width=0.18\linewidth]{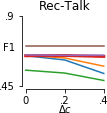}\quad
		\includegraphics[width=0.18\linewidth]{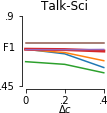}\quad
		\includegraphics[width=0.18\linewidth]{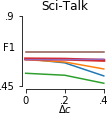}\quad
		\vspace{5pt}
		
		\includegraphics[width=0.7\linewidth]{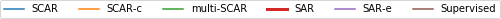}
		
	}
\end{figure}

\subsubsection{Results}
\label{sec:scarres}
SAR-EM performs very well at estimating the class prior, as can be seen in \figureref{fig:c} and \tableref{tab:c}. It has the best rank and lowest absolute error on average.

\subsection{Performance under the SAR Assumption}

The 20 newsgroups datasets are used to simulate SAR datasets. They always consider two categories to classify. With four categories there are six combinations and 12 datasets in total by switching which category is the positive class.

The attributes of 20 newsgroups are all binary: they indicate if a word appears in the article or not. To select propensity attributes that are sure to have an impact on the labeling, we only considered the attributes to have a frequency between 30\% and 70\% in the data, which leaves between five and eight attributes per dataset.

We consider two types of propensity scores. The first one only depends on one attribute. We analyze how deviating from the SCAR setting affects learners that assume SCAR. To this end, the difference between the two propensity scores $\Delta c$ is varied with steps of $0.2$, centering them around $\overline{c}=0.3$ or $\overline{c}=0.5$:
\begin{align*}
e(x_e) = x_e(\overline{c}~+\hspace{-3pt}/\hspace{-3pt}-~\Delta c/2) + (1-x_e)(\overline{c}~-\hspace{-3pt}/\hspace{-3pt}+~\Delta c/2)
\end{align*}

The second type of propensity score is based on three attributes, where all attributes contribute independently: 0.9 for the attribute being 1 and 0.5 otherwise:
\begin{align*}
e(x_e) = 0.9^{\sum x_e}\cdot 0.5^{3-\sum x_e}
\end{align*}

The labels were generated 
%five times 
for each combination of attributes, at random according to the propensity scores functions.

The following methods are compared:
\begin{description}[noitemsep,nolistsep]
	\item[Supervised] Logistic regression with all class labels available.
	\item[SAR] SAR-EM as described in this paper.	
	\item[SAR-e] SAR-EM with access to the true propensity score (\sectionref{sec:sar-e}).
	\item[SCAR] SAR-EM is used for the SCAR setting, as it outperformed the others (\sectionref{sec:scarres}).
	\item[multi-SCAR] Independent SCAR models for all propensity attributes configurations.
	\item[SCAR-c] SAR-EM with access to a propensity score which is the true label frequency.
\end{description}

We compare the algorithms based on their classification F1 score and propensity score accuracy. The datasets were randomly divided into five folds where four were used for learning and one for evaluation. The folds are assigned in five different ways.

\subsubsection{Results}

As expected, the more the dataset deviates from the SCAR assumption, the worse the results get when this is assumed. This result supports the need for methods like SAR-EM that do not make this strong assumption (\figureref{fig:delta3,fig:delta5}).

Interestingly, learning with an unknown propensity score function performs almost equally well as with a known one, even when multiple attributes are involved (\tableref{tab:acc}).

Knowing the label frequency for the SCAR assumption does help when the assumption does not hold. This is due to the label frequency always being an overestimate when the assumption does not hold as a result of the local certainty assumption.

\fix{
\subsection{Effect of Unbalanced Classes}
Unbalanced classes are simulated for the datasets with three propensity features. To this end, either the positive or negative class was subsampled to 30\%, assuming completely balanced classes this results in class priors $\alpha = 0.23$ and $\alpha = 0.77$. 
}
\fix{
\subsubsection{Results}
\figureref{fig:unbalanced} compares the F1 scores of the trained classifiers for the different class priors. Handling unbalanced domains clearly becomes harder with PU data. Knowing the propensity score (SAR-e) or the class prior (SCAR-c) gives an advantage here.
}

\begin{figure}[tbp]
	\floatconts
	{fig:delta5}
	{\caption{Propensity score with one variable, centered around $\overline{c}=0.5$. Assuming SCAR ($\Delta c=0$) hurts the performance when the reality deviates from it.}}
	{%
		
		\includegraphics[width=0.18\linewidth]{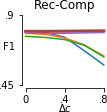}\quad
		\includegraphics[width=0.18\linewidth]{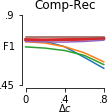}\quad
		\includegraphics[width=0.18\linewidth]{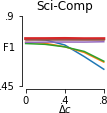}\quad
		\includegraphics[width=0.18\linewidth]{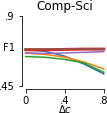}\quad
		\vspace{10pt}
		
		\includegraphics[width=0.18\linewidth]{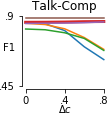}\quad
		\includegraphics[width=0.18\linewidth]{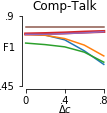}\quad
		\includegraphics[width=0.18\linewidth]{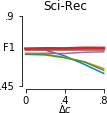}\quad
		\includegraphics[width=0.18\linewidth]{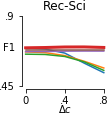}\quad	
		\vspace{10pt}
		
		\includegraphics[width=0.18\linewidth]{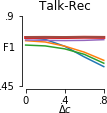}\quad
		\includegraphics[width=0.18\linewidth]{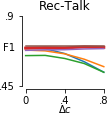}\quad
		\includegraphics[width=0.18\linewidth]{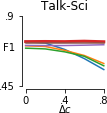}\quad
		\includegraphics[width=0.18\linewidth]{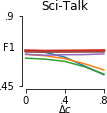}\quad
		\vspace{5pt}
		
		\includegraphics[width=0.7\linewidth]{figures/cd_exp_legend}
		
	}
	
\end{figure}

\begin{figure}[tbp]
	\floatconts
	{fig:unbalanced}
	{\caption{\fix{Unbalanced domains are relatively harder when learning from positive and unlabeled data than from fully supervised data. Knowing the true class prior (SCAR-c and SAR-e) gives an advantage.}}}
	{%
		\includegraphics[width=0.18\linewidth]{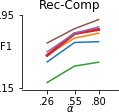}\quad
		\includegraphics[width=0.18\linewidth]{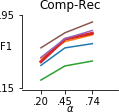}\quad
		\includegraphics[width=0.18\linewidth]{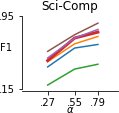}\quad
		\includegraphics[width=0.18\linewidth]{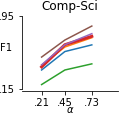}\quad
		\vspace{10pt}
		
		\includegraphics[width=0.18\linewidth]{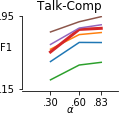}\quad
		\includegraphics[width=0.18\linewidth]{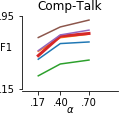}\quad
		\includegraphics[width=0.18\linewidth]{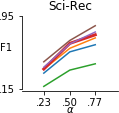}\quad
		\includegraphics[width=0.18\linewidth]{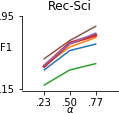}\quad	
		\vspace{10pt}
		
		\includegraphics[width=0.18\linewidth]{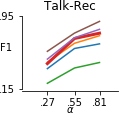}\quad
		\includegraphics[width=0.18\linewidth]{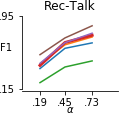}\quad
		\includegraphics[width=0.18\linewidth]{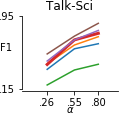}\quad
		\includegraphics[width=0.18\linewidth]{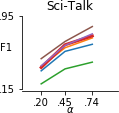}\quad
		\vspace{5pt}
		
		\includegraphics[width=0.7\linewidth]{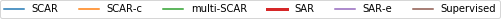}
		
	}
\end{figure}

\begin{table}[tbp]
	\floatconts
	{tab:acc}%
	{\caption{Classifier \fix{F1 score} (left) and propensity (right) score accuracies. `Supervised' has no propensity model. SAR-EM with the SAR assumption performs \fix{similar to} learning with a known propensity score function. The SCAR assumption is clearly insufficient. }}%
   {\small 
   	\begin{tabular}{l|cc|cccc}
   		Data&Supervised&SAR-e&SAR&multi-SCAR&SCAR&SCAR-c\\\hline
   		\fix{Sci-Rec}&\fix{0.68}&\fix{0.64\quad0.65}&\fix{{\bf 0.65}\quad{\bf 0.64}}&\fix{0.36\quad0.29}&\fix{0.56\quad0.54}&\fix{0.60\quad0.59}\\
   		\fix{Rec-Sci}&\fix{0.68}&\fix{0.64\quad0.64}&\fix{{\bf 0.65}\quad{\bf 0.63}}&\fix{0.36\quad0.30}&\fix{0.57\quad0.53}&\fix{0.61\quad0.58}\\
   		\fix{Talk-Rec}&\fix{0.76}&\fix{0.71\quad0.65}&\fix{{\bf 0.70}\quad{\bf 0.63}}&\fix{0.38\quad0.29}&\fix{0.59\quad0.53}&\fix{0.65\quad0.59}\\
   		\fix{Rec-Talk}&\fix{0.71}&\fix{0.66\quad0.62}&\fix{{\bf 0.66}\quad{\bf 0.61}}&\fix{0.39\quad0.34}&\fix{0.60\quad0.51}&\fix{0.63\quad0.55}\\
   		\fix{Sci-Comp}&\fix{0.74}&\fix{0.71\quad0.67}&\fix{{\bf 0.71}\quad{\bf 0.66}}&\fix{0.37\quad0.25}&\fix{0.60\quad0.58}&\fix{0.65\quad0.63}\\
   		\fix{Comp-Sci}&\fix{0.69}&\fix{0.64\quad0.64}&\fix{{\bf 0.64}\quad{\bf 0.63}}&\fix{0.36\quad0.30}&\fix{0.56\quad0.52}&\fix{0.61\quad0.58}\\
   		\fix{Rec-Comp}&\fix{0.80}&\fix{0.75\quad0.67}&\fix{{\bf 0.75}\quad{\bf 0.65}}&\fix{0.39\quad0.25}&\fix{0.65\quad0.57}&\fix{0.70\quad0.62}\\
   		\fix{Comp-Rec}&\fix{0.75}&\fix{0.69\quad0.64}&\fix{{\bf 0.68}\quad{\bf 0.63}}&\fix{0.39\quad0.30}&\fix{0.59\quad0.51}&\fix{0.66\quad0.58}\\
   		\fix{Talk-Comp}&\fix{0.89}&\fix{0.82\quad0.67}&\fix{{\bf 0.80}\quad{\bf 0.65}}&\fix{0.41\quad0.25}&\fix{0.66\quad0.55}&\fix{0.75\quad0.63}\\
   		\fix{Comp-Talk}&\fix{0.83}&\fix{0.74\quad0.63}&\fix{{\bf 0.73}\quad{\bf 0.61}}&\fix{0.43\quad0.33}&\fix{0.65\quad0.50}&\fix{0.71\quad0.56}\\
   		\fix{Talk-Sci}&\fix{0.73}&\fix{0.69\quad0.65}&\fix{{\bf 0.69}\quad{\bf 0.64}}&\fix{0.36\quad0.28}&\fix{0.59\quad0.55}&\fix{0.63\quad0.60}\\
   		\fix{Sci-Talk}&\fix{0.67}&\fix{0.63\quad0.63}&\fix{{\bf 0.64}\quad{\bf 0.62}}&\fix{0.37\quad0.33}&\fix{0.56\quad0.51}&\fix{0.60\quad0.56}\\
   	\end{tabular}
   	}
\end{table}

\section{Conclusions}
\label{sec:concl}
This paper considers learning from positive and unlabeled data (PU learning) under the Selected At Random (SAR) assumption. That assumption states that the probability of selecting a positive example to be labeled depends on the attributes, in contrast to being constant, as commonly assumed in PU learning under the Selected Completely At Random (SCAR) assumption. The SCAR assumption is clearly often violated in practice and our experiments show that using it when it does not hold severely hurts the performance.

In this work, we analyzed common assumptions in PU learning and investigated how they can be used under the SAR assumption, by formulating the problem as a multiple problems under SCAR assumption. This results in a simple yet effective EM-based method. In our experiments, we show that this method is very promising, as it does virtually equally well as assuming that the labeling mechanism is known.

%Future work
Future work will investigate if the method is still effective for more complex propensity score functions. More powerful classifiers could then be preferred, but they might need calibration of their output probabilities.

\section*{Acknowledgements}
JB is supported by IWT (SB/141744).  JD is partially supported by KU Leuven Research Fund (C14/17/070 and C22/15/015), FWO-Vlaanderen (SBO-150033, EOS No. 30992574) and Interreg V A project NANO4Sports.

\bibliography{Bekker-LIDTA18}

\appendix

\end{document}